\documentclass[times,10pt,twocolumn]{article}
\usepackage{latex8}
\usepackage{times}
\usepackage{balance}

\AtBeginDocument{%
  \providecommand\BibTeX{{%
    \normalfont B\kern-0.5em{\scshape i\kern-0.25em b}\kern-0.8em\TeX}}}

\usepackage{xspace}
\usepackage{caption}
\usepackage[skip=0cm,list=true,labelfont=bf]{subcaption}

\usepackage{soul}
\usepackage{lipsum,graphicx,multicol}
\usepackage{tabularx}

\usepackage{color, colortbl}
\definecolor{Gray}{gray}{0.9}
\usepackage[first=0,last=9]{lcg}

\usepackage[]{graphicx} 

\usepackage{algorithm}
\usepackage[noend]{algpseudocode}
\usepackage{nicefrac}
\providecommand{\keywords}[1]{\textbf{\textit{Index terms---}} #1}

\usepackage{amsmath}
\usepackage{amssymb}
\usepackage{algorithm}
\usepackage[noend]{algpseudocode}

\usepackage{booktabs}
\usepackage{comment}

\makeatletter
\def\BState{\State\hskip-\ALG@thistlm}
\makeatother

\begin{document}

\title{Bi-GRU Based Deception Detection using EEG
 Signals}
\author{Danilo Avola, Muhammad Yasir Bilal, Emad Emam, Cristina Lakasz, \\ Daniele Pannone, and Amedeo Ranaldi\\
Department of Computer Science, Sapienza University of Rome\\
Via Salaria 113, 00198, Rome (RM), Italy\\
\{avola,bilal,emam,lakasz,pannone,ranaldi\}@di.uniroma1.it\\
}

\maketitle
\thispagestyle{empty}
\begin{abstract}
Deception detection is a significant challenge in fields such as security, psychology, and forensics. This study presents a deep learning approach for classifying deceptive and truthful behavior using ElectroEncephaloGram (EEG) signals from the Bag-of-Lies dataset, a multimodal corpus designed for naturalistic, casual deception scenarios. A Bidirectional Gated Recurrent Unit (Bi-GRU) neural network was trained to perform binary classification of EEG samples. The model achieved a test accuracy of 97\%, along with high precision, recall, and F1-scores across both classes. These results demonstrate the effectiveness of using bidirectional temporal modeling for EEG-based deception detection and suggest potential for real-time applications and future exploration of advanced neural architectures.
 \end{abstract}


\section{Introduction}
Computer vision and machine learning have emerged as pivotal technologies enabling a wide range of applications, including robust feature extraction \cite{Avola2018features,guangyun2021orb}, person re-identification in surveillance systems \cite{avola2020bodyprint,kunho2023reid}, and intelligent drone navigation \cite{Avola2021,xu2024uav}. Moreover, recent advancements extend their impact to more sensitive domains such as lie detection \cite{avola2019automatic,avola2020lie}, demonstrating their growing role in addressing complex real-world challenges. Deception detection refers to the process of identifying statements or actions that are designed to conceal the truth. Lying is a common habitual practice, intrinsic in people daily life. It is characterized by an intentional effort to mislead others for personal benefit or gain and is a behavior that can manifest in various forms, such as fabrications, omissions, and misrepresentations. 
Fabrications involve the creation of false information or stories, omissions refer to the deliberate exclusion of certain truths, and misrepresentations involve presenting information in a way that distorts the truth or creates a false impression. The spectrum of deception is thus broad, ranging from harmless white lies to serious threats that pose significant morality and security risks to society, with far-reaching implications, affecting interpersonal relationships, organizational dynamics, and even national security. For these reasons, the importance of deception detection cannot be overstated, as it plays a crucial role in various aspects of society, including law enforcement and corporate governance. By accurately identifying deceptive behavior, we can foster transparency, accountability, and trust in our interactions. In recent studies \cite{prome2024ldnet}, the methods for detecting deception have been broadly divided into two main categories: non-contact based and contact based methods. The first primarily focuses on the extraction of biological and physiological data that measure activity changes within the body, such as heart rate, blood pressure, respiratory patterns and, most importantly for our study, bioelectrical signals. The second group includes methods that rely on acoustic, visual, thermal data, and verbal communication. Acoustic data can provide insights into changes in a person’s voice, while verbal data can reveal deceptive cues in the content of a person’s speech. Thermal data can show physiological changes such as increased body temperature, and visual data, such as facial expressions or body language, can provide clues about whether a person is telling the truth or attempting to deceive \cite{defofdeceiving}. By exploiting modern deep learning techniques, this paper introduces a novel application of a Bidirectional Gated Recurrent Unit (Bi-GRU) neural network architecture for classifying truthful and deceptive behavior based solely on EEG signals, specifically focusing on the temporal dynamics of brain activity. The following are the contributions of our work:
\begin{itemize}
    \item \textbf{Exclusive Focus on Spontaneous Deception:} Utilizes the EEG portion of the Bag-of-Lies dataset, which captures spontaneous, self-initiated deception in realistic scenarios, unlike many prior datasets relying on forced or artificial deception;
    \item \textbf{Robust pre-processing and augmentation pipeline:} Proposes a structured pre-processing method that includes filtering, overlapping window segmentation, class balancing via undersampling, and data augmentation by Gaussian noise to address dataset limitations;
    \item \textbf{High Classification Accuracy and Generalization:} Achieves a 97\% accuracy on the EEG classification task, outperforming several state-of-the-art models applied to the same dataset, including Random Forest, Bi-LSTM, and CNN-based approaches.
\end{itemize}

The rest of the paper is organized as follows. Section \ref{sec:related_work}, the current state-of-the-art methods in deception detection through EEG are discussed. Section \ref{sec:method} presents the designed method, together with the description of the pre-processing operations performed on the dataset. Section \ref{sec:evalutation} report the conducted experiments and the results obtained. Finally, Section \ref{sec:conclusion} concludes the paper.

\section{Related work} \label{sec:related_work}
The ability to detect lies is of critical importance in legal and security settings, for this reason research on deception detection has been extensively conducted. 
With advances in technology, particularly in the field of neuroscience and artificial intelligence, new methods have emerged to improve the accuracy and reliability of lie detection. We will particularly focus on EEG applications in this field. Much of the work related to deception detection using EEG data is focused specifically on P300 waves. These are components of the Event-Related Potential (ERP) characterized by a positive deflection in voltage that occurs approximately 300 milliseconds after the presentation of a stimulus and is often more pronounced when a subject recognizes the stimulus as familiar or meaningful. We now discuss briefly some works that inspired our study, but that focus on less controlled experiments, closer to real life scenarios and that do not leverage P300 waves. 

Authors in \cite{lr10} present the Bag-Of-Lies dataset a pivotal multimodal dataset in this field. It integrates video, audio, EEG, and gaze data, collected from 35 unique subjects in a realistic setting. It provides a balanced set of 325 annotated data points divided almost equally between truthful and deceptive responses. The video data is extracted using Local Binary Patterns (LBP) and analyzed thru SVM, Random Forest, and MLP. Extracted features from the audio data, such as spectral properties and mel-frequency cepstral coefficients (MFCCs), are classified using Random Forest and KNN. EEG data is analyzed through sub-sampling using a sliding window approach, with features fed into a CNN architecture and Random Forest for classification. Finally, gaze data features are calculated included fixations, eye blinks, and pupil size, which are then used to construct a feature vector for machine learning analysis. Individual modal analyses showed varied effectiveness, the EEG data provided moderate results, with the best classification accuracy using Random Forest achieving 58.71\%. Audio data achieved 56.20\% accuracy with Random Forest. Similarly to video, audio data provided modest results with the best performance shown by KNN at 56.22\% accuracy. Gaze data performed better than all other modalities alone, with Random Forest achieving the highest accuracy at 61.70\%. Performing multimodal fusion classification, particularly combining gaze, video, and audio data, yielded the highest improvements in detection accuracy, 64.69\%. Lastly, the combination of all four modalities (Video, Audio, EEG, and Gaze) resulted in the highest accuracy of 66.17\% highlighting the benefit of a comprehensive analytical approach. Furthermore, the EEG data of the Bag-Of-Lies dataset is the one used in this study. For this reason, part of this dataset, will be explained in greater detail in the following chapters. The work presented in \cite{lr11} compares a deep convolutional neural network (DCNN) designed for detecting deception on three multimodal databases: Bag-of-Lies (BoL), Real-Life (RL) trail, and Miami University Deception Detection (MU3D). Visual, vocal, and EEG signals of BoL are considered. Whereas for the other two databases, only vocal and visual modalities are available. To analyze visual data, twenty frames from each video are selected, cropped to focus on facial regions, and resized to 256x256 pixels. These frames are concatenated to form a single image and further resized to 256x1024 pixels. Audio signals are extracted from videos and plotted into a 2-D plane, creating images of 256x256 pixels. Finally, thirteen channels of EEG signals are also plotted into 2-D planes, concatenated to form a single image of 256x3328 pixels, and resized to 256x1024 pixels. These three data inputs are given to the LieNet model that employs multiple convolutional layers with various kernel sizes (1x1, 3x3, 5x5, and 7x7) to capture multiscale features from the input images. The network architecture includes nine convolutional layers, four max-pooling layers, and two fully connected layers, it uses ReLU activation functions for nonlinearity and a softmax classifier to estimate scores. Scores from each modality are combined using a score level fusion technique, assigning weights to each modality based on its performance, so to ensure that the most informative modalities contribute more significantly to the final classification decision. The LiNet approach obtained an average accuracy of 95.91\% on the BoL dataset, a 97.33\% accuracy on the RL Trail database, and lastly 98.14\% accuracy on MU3D. The work presented in \cite{lr7} utilizes the Bag-of-Lies dataset, specifically its EEG data, and develops a model combining Long Short-Term Memory (LSTM) and Neural Circuit Policies (NCP). The data were processed using Discrete Wavelet Transform (DWT) to decompose the EEG signals into sub-signals. For classification the model used a combination of LSTM, that handles sequence prediction problems, and NCP, that reduces the network complexity and computational load. The model performance achieved accuracy of 97.88\%. Authors in \cite{lr13} present a model utilizing the Bag-Of-Lies dataset. Dense optical flow features are extracted from video frames to analyze spatial and temporal features of facial movements thru two-stream convolutional neural network (CNN). Audio signals are transformed into frequency-distributed spectrograms, and an attention-augmented CNN is employed to capture variations in speech patterns.
EEG signals are treated as time series data and bi-directional long short-term memory (Bi-LSTM) networks are used to analyze them.
The combined system achieved an accuracy of 83.5\%. Individual modal analyses also showed promising results, with EEG-based analysis providing the highest accuracy 82.24\% among the single modalities. Table \ref{tab:literaturereview} provides a summary of the literature review presented.

\begin{table}[t]
\centering
\caption{State of the Art EEG Classification for Deception Detection.}
\begin{tabular}{lcc}
\hline
\hline
\multicolumn{3}{c}{\textbf{Bag-of-Lies (BoL) Dataset Approaches}} \\
\hline
\textbf{Study} & \textbf{Method} & \textbf{Accuracy (\%)}
\\
\midrule
\cite{lr10} & Random Forest & 58.71 \\
\cite{lr11} & DCNN (LieNet) & 95.91 \\
\cite{lr7} & LSTM + NCP & 97.88 \\
\cite{lr13} & Bi-LSTM & 82.24 \\
\hline
\hline
\end{tabular}
\label{tab:literaturereview}
\end{table}

\section{Method}
\label{sec:method}
This section provides the details of the proposed method, providing details on the pre-processing steps performed on data and on the model architecture.

\subsection{Data Pre-processing}
For each subject $s \in \mathcal{S}$, where $ \mathcal{S} $ denotes the set of all participants, only the EEG signal data are retained. Metadata unrelated to the EEG measurements, such as channel quality indicators and headset positioning parameters, are excluded from the analysis. All recording sessions corresponding to each subject $ s $ and each run $ r \in \mathcal{R}_s $, where $ \mathcal{R}_s $ denotes the set of recording sessions for subject $ s $, are aggregated and annotated. These sessions are concatenated to form a unified dataset $ \mathcal{D} \in \mathbb{R}^{n \times d} $, where $ n $ is the total number of time samples across all subjects and sessions, and $ d $ is the number of retained features. The resulting DataFrame contains the following columns:
\begin{itemize}
    \item $ \mathbf{X}_i \in \mathbb{R}^c $: EEG channel values for $ c $ electrodes,
    \item $ u_i \in \mathcal{S} $: subject identifier;
    \item $ r_i \in \mathcal{R}_s $: run/session identifier;
    \item $ y_i \in \mathcal{Y} $: corresponding annotation label for supervised learning tasks.
\end{itemize}

Any rows with missing or corrupted values are removed to ensure data integrity, resulting in a cleaned dataset $ \mathcal{D}' \subseteq \mathcal{D} $.

To isolate task-relevant neural activity and suppress both low-frequency drift and high-frequency noise, a band-pass filter is applied to each EEG channel signal. Specifically, a high-pass filter with a cut-off frequency $ f_{\text{low}} = 1~\mathrm{Hz} $ and a low-pass filter with $ f_{\text{high}} = 30~\mathrm{Hz} $ are used, so that the filtered signal $ \tilde{\mathbf{X}}_i $ satisfies:
\begin{equation}
\tilde{\mathbf{X}}_i = \mathcal{F}_{[1,30]}(\mathbf{X}_i),
\end{equation}

\noindent where $ \mathcal{F}_{[1,30]} $ denotes the band-pass filtering operator with passband $[1, 30]~\mathrm{Hz}$. This preprocessing step ensures that the preserved EEG components correspond primarily to cognitive and sensory processes of interest, while minimizing the influence of artifacts such as eye blinks and muscle movements. After applying the band-pass filter to the EEG signals, the resulting time series are segmented into overlapping windows to create individual training samples. Each window comprises $ T = 64 $ consecutive time points, and a sliding window approach is adopted with a stride of $ sr = 32 $ time points between consecutive windows. This overlapping strategy increases the number of available samples and ensures temporal continuity, reducing the risk of losing relevant transient neural patterns between non-overlapping segments. Formally, given a filtered EEG signal $ \tilde{\mathbf{X}} \in \mathbb{R}^{n \times c} $, where $ n $ is the number of time points and $ c $ the number of channels, the windowed segments are defined as:

\begin{align}
 \tilde{\mathbf{X}}^{(j)} = \tilde{\mathbf{X}}[jsr : jsr + T - 1], \\ \quad j = 0, 1, \ldots \left\lfloor \frac{n - T}{sr} \right\rfloor.
\end{align}

Each segment $ \tilde{\mathbf{X}}^{(j)} \in \mathbb{R}^{T \times c} $ is associated with a label $ y^{(j)} \in \mathcal{Y} $ inherited from the original recording. Given the dataset exhibits class imbalance, with $ 108 $ samples labeled as ``truth'' and $ 93 $ as ``lie'', a balancing procedure is applied to avoid classifier bias toward the majority class. Specifically, undersampling is performed by randomly removing samples from the over-represented class until both classes contain an equal number of instances. Let $ N_1 $ and $ N_2 $ denote the number of samples in the majority and minority classes, respectively (assuming $ N_1 > N_2 $). The balanced dataset $ \mathcal{D}_b $ then satisfies:

\begin{equation}
    |\mathcal{D}_b^{(1)}| = |\mathcal{D}_b^{(2)}| = \min(N_1, N_2),
\end{equation}

\noindent where $ \mathcal{D}_b^{(1)} $ and $ \mathcal{D}_b^{(2)} $ represent the subsets corresponding to the two classes (truth and lie, respectively).
\begin{figure*}[t]
    \centering
    \includegraphics[width=0.6\linewidth]{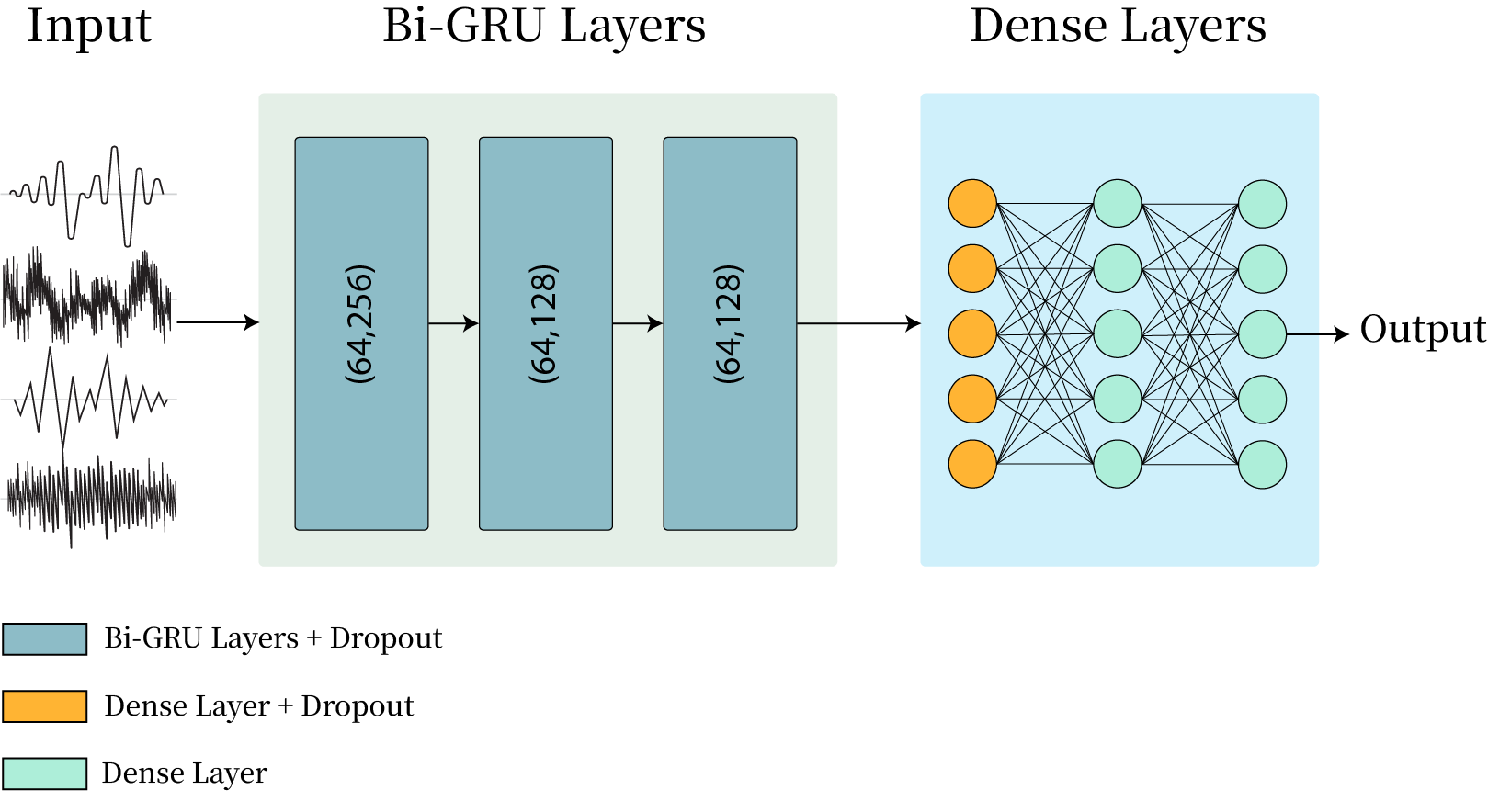}
    \caption{Architecture of the proposed Bi-GRU-based model for EEG signal classification. The input consists of multichannel EEG time series data structured as a tensor of shape (64,64,13)(64,64,13), where 64 is the batch size, 64 is the number of time steps, and 13 denotes the EEG channels. The sequence is processed through three stacked Bi-GRU layers with dropout regularization, producing hidden representations of size (64,256)(64,256), (64,128)(64,128), and (64,128)(64,128), respectively. These representations are then fed into a series of fully connected dense layers, including dropout layers for regularization, to compute the final output probabilities. Color coding distinguishes Bi-GRU layers with dropout (blue), dense layers with dropout (orange), and standard dense layers (green).}
    \label{fig:proposed_model}
\end{figure*}
In order to perform data augmentation, and improve generalization, Gaussian noise, with a noise factor of 0.02, is added to the samples to generate additional synthetic samples. All samples are combined and separated into two sets of training and testing data (80\% for training and 20\% for testing) to build the model and evaluate its performance on previously unseen data.

\subsection{Model Architecture}
The novel method proposed is a BiGRU neural network, an implementation of Bidirectional Gate Recurrent Unit Neural Network \cite{birnn}, with the goal of conducting binary classification on EEG data to detect instances of deception in the recorded subject. In Figure \ref{fig:proposed_model}, a visual representation of the proposed method is depicted.

The input layer is the initial layer of the model, responsible for defining the shape of the input tensor that represents the multichannel EEG time series data. Let the input tensor be denoted as $\mathbf{X} \in \mathbb{R}^{B \times T \times F}$, where $B$ is the batch size (number of samples per batch), $T$ is the number of time steps (i.e., the temporal length of each sample), and $F$ is the number of features (i.e., EEG channels).

In our implementation, we fix the batch size to $ B = 64 $, a common choice in deep learning that balances memory efficiency and convergence speed. Each sample in the batch consists of a temporal window of $ T = 64 $ steps, reflecting the sampling window used in preprocessing. The EEG signal is recorded from $ F = 13 $ channels, which corresponds to the number of spatial sensors (electrodes) deployed.

Therefore, the input tensor for each training batch is formally represented as
$\mathbf{X} \in \mathbb{R}^{64 \times 64 \times 13}$. This structured representation allows the model to capture both temporal dynamics and spatial relationships among the EEG channels across the batch of input sequences. The input tensor is given to the Bidirectional model in order to extract temporal features and learn dependencies across the sequence. The model architecture relies on three Bi-GRU layers, that process information recurrently, both in the forward and backward directions. Each of these layers is followed by a dropout layer set at a 0.5 rate, to randomly ignore half of the neurons during each training step. This is done for the purpose of regularization to prevent overfitting. The first BiGRU layer contains 128 units, or neurons, in each direction, for a total of 256 hidden states, each of which is able to learn features from the data. The second layer is composed of 64 units, for a total of 128 hidden states. The number of neurons is reduced in order to capture and maintain relevant information only. Finally, the third and last recurrent layer is made of 32 units for a total of 64 for both directions. Three dense layers are then applied to perform classification on the extracted features. These are fully connected layers of 64 and 32 neurons, each of which computes a linear transformation on the data received from the previous layer. To introduce non linearity, Rectified Linear Unit (ReLU) is used as activation function. Each of these two layers is followed by a Dropout layer with probability 0.5 as well. Ultimately, the third dense layer, i.e., the last layer of the architecture, contains only 2 units, and softmax activation function is used to output the probability distribution for the input to be of the two classes, truth or lie. 

\section{Evaluation}
\label{sec:evalutation}
In this section, the evaluation of the model, together with details of the used dataset, the training process, and comparison with state-of-the-art approaches, are provided.

\subsection{Dataset}
\begin{figure}[t]
    \centering
    \includegraphics[width=\linewidth]{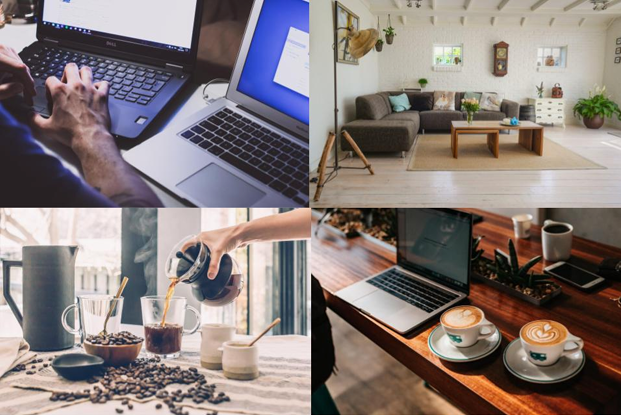}
    \caption{Examples of images used as stimuli during data collection.}
    \label{fig:BoL}
\end{figure}

\begin{figure}[t]
    \centering
    \includegraphics[width=0.7\linewidth]{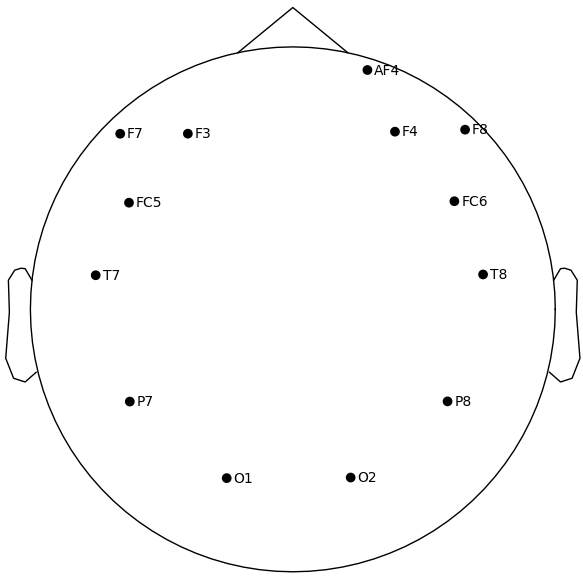}
    \caption{Visual representation of the 13 EEG channels used in the Bag-of-Lies dataset.}
    \label{fig:electrodesplacement}
\end{figure}

The dataset used in this study is the Bag-of-Lies dataset \cite{lr10}. This is a multimodal dataset tailored for the task of deception detection. It includes four types of of data modalities: video, audio, gaze, and EEG data,
 registered from 35 unique subjects. It contains 325 manually annotated recordings of variable length (from 3.5 to 42 seconds), evenly distributed across truths (163) and lies (162). Data was recorded using a smartphone camera and microphone for video and audio. For EEG data, a 14-Channel Emotiv EPOC+ EEG headset was used, and for gaze data the Gazepoint GP3 Eye Tracker. Participants were fitted with an EEG headset and were then seated in front of a monitor equipped with the Eye Gaze sensor and the smartphone. Each participant was shown between 6 to 10 images, one at a time, from a pre-selected set of 21 unique, content-heavy, and descriptive images, chosen to provide rich visual content, making it possible for participants to describe them in various ways. In Figure \ref{fig:BoL}, example images used as stimuli are shown. The key aspect of the task was that participants could choose to describe the image either truthfully or deceptively, creating a natural and spontaneous setting for deception detection.
 Videos aimed at record the facial expressions and body language of the subjects, audio recordings aimed at capturing speech and vocal characteristics, gaze data was used to track eye movements, fixations, blinks, and pupil size, and EEG data was collected to monitor brain activity. Unfortunately, due to thick hair, the Emotiv headset did not fit some participants, for whom EEG data was not possible to be recorded. For this reason, the dataset can be divided in two sets, set A contains the data of the 22 unique users for whom all four data modalities are available, and set B contains gaze, video and audio recordings of all 35 users. EEG data is the only data modality of this dataset that we employ in this study, in order to investigate the cognitive aspects related to deception. We then considered only set A of the dataset, consisting of 201 recordings, 108 of which are truths and 93 are lies. Although the Emotiv headset used consisted of 14 channels, only 13 could be used due to driver issues on channel AF3, the position of available channels (available channels- F3,
 FC5, F7, T7, P7, O1, O2, P8, T8, F8, AF4, FC6, F4) is depicted in Figure \ref{fig:electrodesplacement}.
 We decided to use the Bag-of-Lies dataset, because this data was collected in a
 realistic scenario where participants were given freedom to choose whether to
 tell the truth or lie when describing an image stimulus. Unlike previous datasets
 that often used hypothetical or forced deception scenarios, this method allows
 for naturalistic and spontaneous deception, which is more reflective of real-world situations.
 Furthermore, publicly available EEG datasets created for deception are scarse, the ones we identified were the Bag-of-Lies and the LieWaves dataset \cite{Aslan2024lie}, that is however focused on ERP and does not capture casual deception. Although the recorded data is of limited size, the Bag-of-Lies dataset best suits the task of deception detection that we want to address.

\subsection{Training}
To train the model the loss function used is binary cross entropy, and the optimization algorithm employed is Adam optimizer. 
Binary cross entropy loss is a suited metric for binary classification tasks and serves the purpose of a quantitative metric for guiding the optimization of the model during training, and to evaluate the model performance during validation.

The optimization algorithm used to update the model parameters is Adam. It keeps track of the first moment (mean) and second moment (variance) of the gradients to adjust the step size during optimization. This allows Adam to handle sparse gradients, efficiently improve convergence speed, and dynamically adapt the learning rate.

The loss and the optimization algorithm are important specifically during backpropagation of the gradients of the loss with respect to each of the model parameters. During training these are in fact computed and updated according to the optimization algorithm. Therefore, for each batch, training consists of forward passing (that in our case includes also backward passing, as the model is bidirectional), during which the model parameters are computed, finally backpropagation using Adam optimizer updates the parameters to minimize the loss. Training is thus an iterative process that occurs in many epochs, where a single epoch is a complete pass through the training data. In our model the maximum number of epochs is 100, each epoch consists of 145 batches, and each batch is made of 64 samples. In order to evaluate how well the model is actually learning, so to check that it is not simply overfitting and remembering data without learning information through it, the validation step during training is always adopted. In particular, 5-fold cross validation is performed. The model is trained 5 times, each time using 4 folds for training and the remaining one for validation. Therefore, training data is split into 5 sets of 20 percent of the training data each. So, each time the model is trained, 80 percent of the training data is always used as previously discussed, and 20 percent is used for the validation phase, where the model parameters are evaluated by making predictions. This is useful when dealing with small datasets like the one in our case, and to prevent overfitting. Each training fold underwent early stopping based on the validation loss to prevent overfitting. The model has been trained for a set maximum of 100 epochs per fold, unless early stopping was triggered if the validation loss did not show improvements in 10 epochs. 
\subsection{Model Performance}
\begin{figure}
    \centering
    \includegraphics[width=\linewidth]{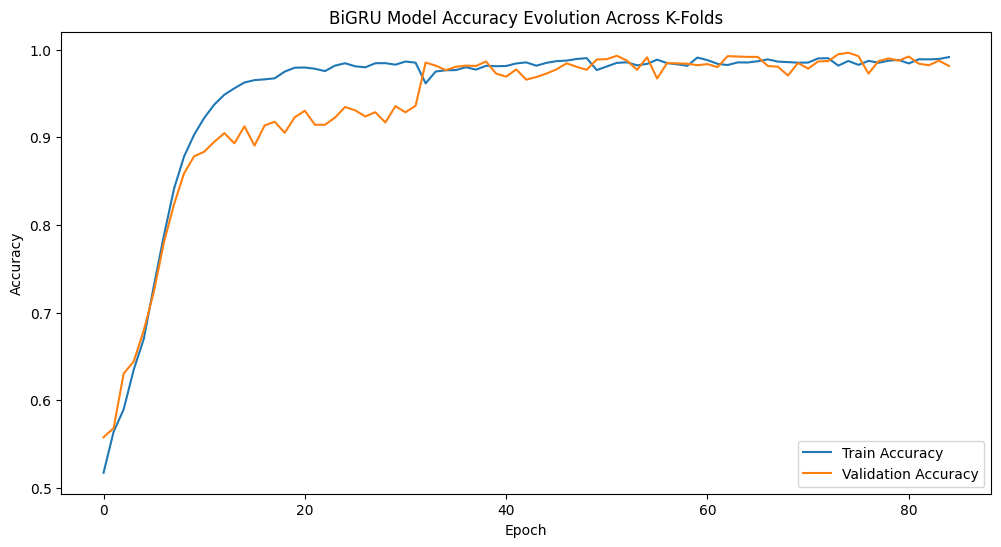}
    \caption{Model Accuracy achieved during training.}
    \label{fig:modelaccuracy}
\end{figure}

\begin{figure}
    \centering
    \includegraphics[width=\linewidth]{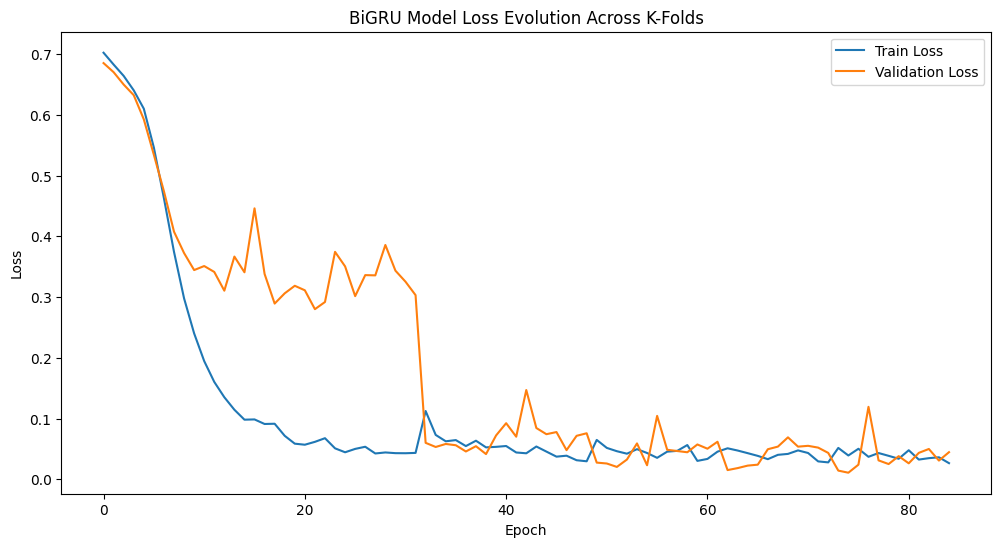}
    \caption{Model loss achieved during training.}
    \label{fig:modelloss}
\end{figure}

\begin{figure}
    \centering
    \includegraphics[width=0.8\linewidth]{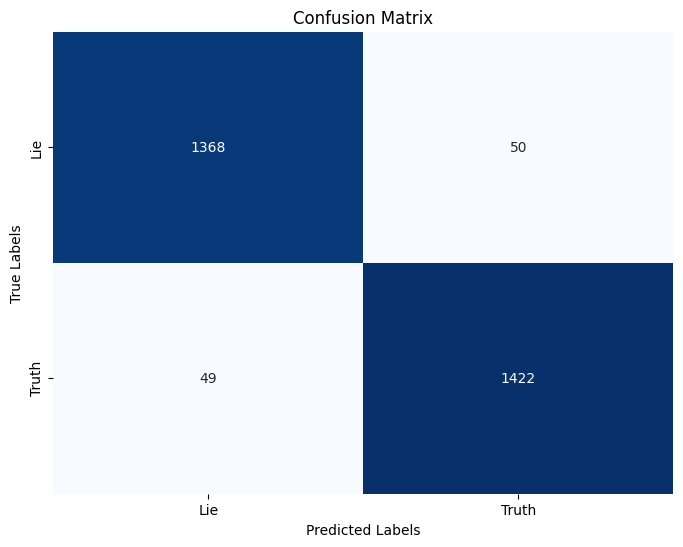}
    \caption{Confusion Matrix related to the performed experiments.}
    \label{fig:confmatrix}
\end{figure}

During training, the model showed consistent improvements, as shown in Figure \ref{fig:modelaccuracy} and Figure \ref{fig:modelloss}. 
After completing training, the model was evaluated on the unseen test set of the data. With a test accuracy of 97\% and a test loss of 0.1696. In order to evaluate the model performance, accuracy, precision, recall and f-1 score have been computed. Accuracy, the percentage of correctly predicted instances, achieved 97\%. Precision, that is the ratio of correctly predicted positive observations to the total predicted positives, achieved 97\% for lies and 97\% for truths. Recall, the ratio of correctly predicted positive observations to all observations in the actual class, achieved 96\% for lies and 97\% for truths. F-1 score, the weighted average of precision and recall, is 97\%. Table \ref{tab:classreport} displays the evaluation metrics of the model. This high accuracy suggests that the model generalizes well to unseen data, successfully classifying the EEG signals with high precision and recall across both classes. Figure \ref{fig:confmatrix} provides the confusion matrix showing the model ability to correctly classify the majority of samples, with only few misclassifications. Out of 1,418 lie samples, the model correctly predicted 1,368 and incorrectly predicted 50. Out of 1,471 true samples, the model correctly predicted 1,422 and incorrectly predicted 49.

\begin{table}[t]
\centering
\caption{Classification report showing precision, recall, f1-score, and support.}
\label{tab:classreport}
\small
\begin{tabular}{lcccc}
\hline
\hline
 & \textbf{Precision} & \textbf{Recall} & \textbf{F1-score} & \textbf{Support} \\
\hline
\textbf{Truth} & 0.97& 0.96& 0.97& 1418 \\
\textbf{Lie} & 0.97 & 0.97& 0.97& 1471 \\
\hline
\textbf{Accuracy} &  &  & 0.97& 2889 \\
\textbf{Macro avg} & 0.97& 0.97& 0.97& 2889 \\
\textbf{Weighted avg} & 0.97& 0.97& 0.97& 2889 \\
\hline
\hline
\end{tabular}
\end{table}

Despite the high performance, some errors were observed in the form of false positives and false negatives, which could be linked to ambiguous signals that may not fit neatly into either class, and outliers or edge cases in the dataset that might not have been fully captured during training. Moreover, the limited size of the dataset could contribute in limiting the generalization capabilities of the model. Additional improvements, such as exploring more advanced signal processing techniques or employing more sophisticated data cleaning procedures, could help reduce these errors. However, the combination of high test accuracy, low loss, and strong precision and recall values across classes demonstrates that the Bi-GRU model effectively generalizes to unseen data and provides reliable predictions for the deception detection classification task on EEG signals, allowing for potential real-life applications. 

We now provide an analysis of this study compared with the current state of the art. In current literature, studies on deception detection using EEG data are scarce. In particular, our study used the Bag-of-Lies dataset, tailored for casual deception scenarios, and proved to be successful in the task of deception detection based on EEG signals, achieving an accuracy of 97\%. Our method is comparable to the state of the art of deception detection, outperforming studies based on similar approaches. 
In particular we now summarize the state of the art of the studies employing the same data as we do. The article that presents the dataset BoL used in this study \cite{lr10} achieves 58.71\% accuracy on EEG classifiaction using Random Forest. The study, LieNet \cite{lr11} converts the EEG data into images and classifies the images using a CNN approach, achieving 95.91\% accuracy. The article \cite{lr7} uses an LSTMNCP approach with accuracy 97.88\%. Lastly, \cite{lr13} exploits Bi-LSTM and achieves 82.24\% accuracy on EEG classification. Our study is inspired by \cite{lr13}, and outperforms it, with an increase of 14\% in accuracy. Furthermore, it introduces a new EEG deception detection method, exploiting a bidirectional GRU architecture, thus contributing in the research of casual deception, whom literature is, as noted, scarce. The results of the studies in the literature conducted on this dataset are summarized again in Table \ref{tab:bolliteraturereview}. 
\begin{table}[t]
\centering
\caption{Accuracy of the state-of-the-art EEG Deception Detection methods applied on Bag-of-Lies dataset.}
\label{tab:bolliteraturereview}
\begin{tabular}{lcc}
\hline
\hline
\textbf{Study} & \textbf{Method} & \textbf{Accuracy (\%)} \\
\hline
\cite{lr10} & Random Forest & 58.71 \\
\cite{lr11} & DCNN (LieNet) & 95.91 \\
\cite{lr7} & LSTM + NCP & 97.88 \\
\cite{lr13} & Bi-LSTM & 82.24 \\
\hline
\hline
\end{tabular}
\end{table}
The study \cite{lr13} is the most similar to our approach, but showed limited performance. The model architecture is simpler and includes one Bi-LSTM layer, data normalized, bandpass filtered, and samples created using a sliding window and zero-padding to ensure the same length.se are then given as input to the model that exploits Bi-LSTM to capture temporal dependencies. The results of this study presented limited classification and generalization capabilities, achieving an average accuracy of 82.4\%.

Our approach shows a considerable increase in performance. By appropriately pre-processing, sampling and augmenting data, and by implementing a more complex Bi-GRU architecture, our model is able to learn and capture key temporal dependencies to make substantially better deception detection predictions on new unseen EEG data. It achieves accuracy of 97\% and outperforms similar state of the art approaches.

\section{Conclusions}
\label{sec:conclusion}
In this study, we explored the use of a Bidirectional Gated Recurrent Unit (Bi-GRU) Neural Network based approach on EEG data for deception detection. Our model achieved classification accuracy of 97\%, supported by robust precision and recall scores for both truth and lie classes. The dataset utilized in this study stands out in the existing literature as it closely mirrors real-world scenarios of deception without relying on Event-Related Potentials (ERP). Unlike many datasets that impose structured experimental settings, this dataset allows participants the freedom to fabricate a deceitful narrative, thereby capturing EEG responses associated with realistic scenarios. The success of the model can be attributed to several factors, including the use of data augmentation techniques, pre-processing of the EEG signals and the application of advanced deep learning architectures such as Bi-GRU. These findings demonstrate the potential of developing deception detection solutions based on EEG signals.

Despite the excellent results, the study highlights several limitations that open avenues for future research. One of the primary challenges was the limited size of the Bag-of-Lies dataset, consisting of only 201 recordings, for about 30 minutes of total recording time, it required data augmentation to mitigate the lack of sufficient samples. Therefore, to improve the generalizability and applicability of such models, a key research direction is the creation of larger datasets, particularly in conditions that more closely mimic real-world scenarios.

In conclusion, this study successfully demonstrated the potential of a Bi-GRU based approach for deception detection using EEG data, achieving high accuracy despite the limitations posed by a small dataset. Moving forward, in order to realize practical applications of EEG-based deception detection systems, future research should address the need for more comprehensive datasets, and explore more sophisticated signal processing and neural network architectures to enhance performance and reliability.

\section*{Acknowledgments}
This work was supported by ``Smart unmannEd AeRial vehiCles for Human likE monitoRing (SEARCHER)'' project of the Italian Ministry of Defence within the PNRM 2020 Program (Grant Number: PNRM a2020.231); ``EYE-FI.AI: going bEYond computEr vision paradigm using wi-FI signals in AI systems'' project of the Italian Ministry of Universities and Research (MUR) within the PRIN 2022 Program (Grant Number: 2022AL45R2) (CUP: B53D23012950001); MICS (Made in Italy – Circular and Sustainable) Extended Partnership and received funding from Next-Generation EU (Italian PNRR – M4 C2, Invest 1.3 – D.D. 1551.11-10-2022, PE00000004) (CUP MICS B53C22004130001); “Enhancing Robotics with Human Attention Mechanism via Brain-Computer Interfaces” Sapienza University Research Projects (Grant Number: RM124190D66C576E).

\balance
\bibliographystyle{latex8}
\bibliography{bibliography}

\end{document}